\documentclass[lettersize,journal]{IEEEtran}
\usepackage[dvipsnames]{xcolor}
\usepackage{amsmath,amsfonts}
\usepackage{algorithmic}
\usepackage{graphicx}
\usepackage{algorithm}
\usepackage{array}
\usepackage{tikz}
\usepackage{comment}
\usepackage{multirow}
\usepackage{tablefootnote}
\usepackage{amsmath}
\usepackage{amssymb}
\usepackage{newfloat}
\usepackage{listings}
\usepackage[caption=false,font=normalsize,labelfont=sf,textfont=sf]{subfig}
\usepackage{textcomp}
\usepackage{stfloats}
\usepackage{url}
\usepackage{verbatim}
\usepackage{graphicx}
\usepackage{bm}
\usepackage{multirow}
\usepackage{booktabs}
\usepackage{pifont}
\usepackage{bbding}
\usepackage{fontawesome}
\usepackage{subfig}
\usepackage{makecell}
\usepackage{subfloat}
\usepackage{multicol}

\newcommand{\lk}[1]{{\color{Green}{#1}}}

\usepackage[pagebackref=false,breaklinks,colorlinks,bookmarks=false]{hyperref}

\makeatletter
\newcommand{\thickhline}{%
    \noalign {\ifnum 0=`}\fi \hrule height 1pt
    \futurelet \reserved@a \@xhline
}

\makeatother


\usepackage{colortbl} 
\definecolor{mypink}{cmyk}{0, 0.7808, 0.4429, 0.1412}
\definecolor{mygreen}{rgb}{0.0, 0.0, 0.0}
\definecolor{myblue}{rgb}{0.0, 0.72, 0.92}
\definecolor{mygray}{gray}{0.6}
\definecolor{mygray-bg}{gray}{0.9}

\usepackage{amssymb}

\newcommand{\lkx}[1]{{\color{Black}{#1}}}

\begin{document}

\title{IDPro: Flexible Interactive Video Object Segmentation by ID-queried Concurrent Propagation}

\author{Kexin Li$^*$,
        Tao Jiang$^*$,
        Zongxin Yang$^\dag$,
        Yi Yang,
        Yueting Zhuang,
        Jun Xiao
\thanks{$^*$ Kexin Li and Tao Jiang are co-first authors.  }
\thanks{$\dag$ Zongxin Yang is the corresponding author.}
\thanks{K. Li, T. Jiang, Y. Yang, Y. Zhuang and J. Xiao are with the College of Computer Science and Technology, Zhejiang University, Hangzhou, China (e-mail: 12221004@zju.edu.cn, 22151142@cs.zju.edu.cn, yangyics@zju.edu.cn, yzhuang@zju.edu.cn, junx@cs.zju.edu.cn). Z. Yang is with the DBMI, HMS, Harvard University, Boston, USA (e-mail: zongxin\_yang@hms.harvard.edu).}
}

\markboth{IEEE TRANSACTIONS ON CIRCUITS AND SYSTEMS FOR VIDEO TECHNOLOGY,~Vol.~xx, No.~xx, FEBRUARY~2023}%
{Shell \MakeLowercase{\textit{et al.}}: Label Semantic Knowledge Distillation for Unbiased Scene Graph Generation}



\maketitle

\begin{figure*}[t]
    \centering
\includegraphics[width=1\textwidth]{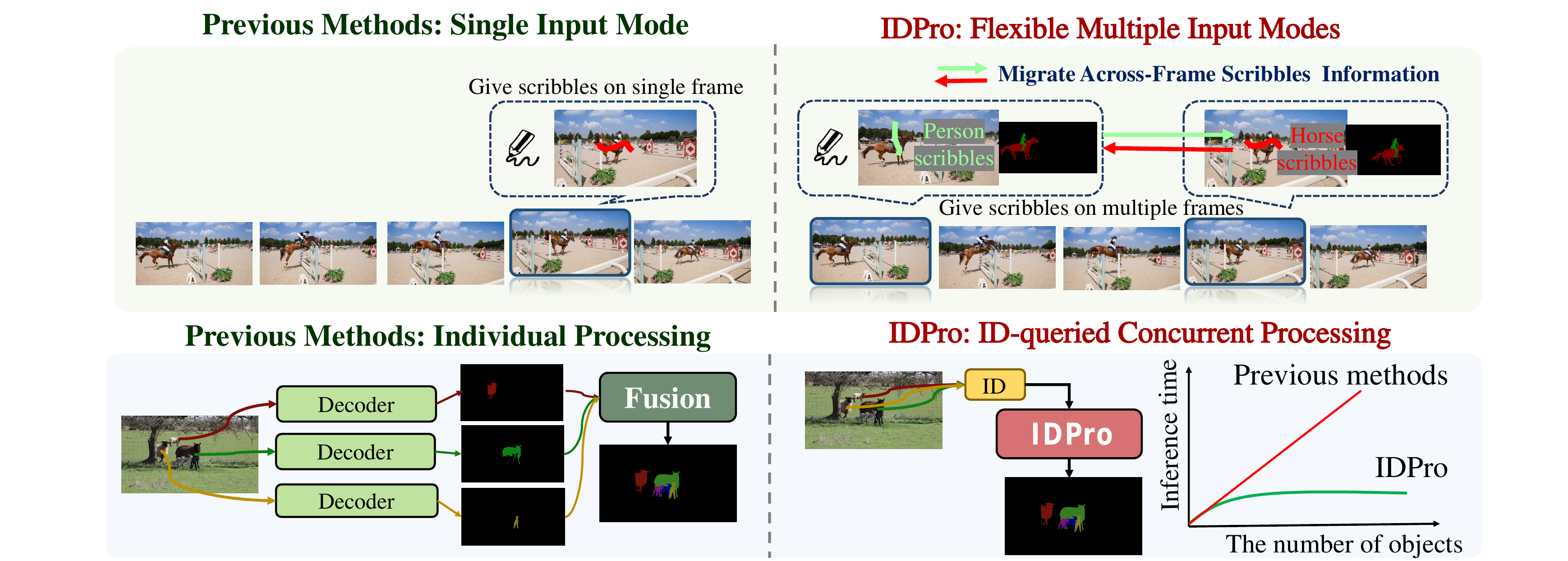}
    \vspace{-0.3cm}
    \caption{IDPro enables users to interact with multiple frames at once, whereas the SOTA method can only accept one single frame. Moreover, IDPro processes multiple objects simultaneously, leading to faster model performance and reduced computing resources.}
    \vspace{-0.3cm}
    \label{fig:introduction}
\end{figure*}

\begin{abstract}

Interactive Video Object Segmentation~(iVOS) is inherently demanding, requiring real-time interaction between humans and computers. Enhancing user experience involves considerations such as user input habits, segmentation quality, running time, and memory consumption. However, existing methods compromise user experience by employing a single input mode and exhibiting slow running speeds. Specifically, these approaches restrict user interaction to a single frame, limiting the expression of user intent. To overcome these limitations and better align with user habits, we introduce a framework that facilitates flexible input modes by ID-queried concurrent propagation~(IDPro). In particular, we have devised the Across-Frame Interaction Module~(AFI), allowing users to freely annotate various objects across multiple frames. The AFI module transfers scribble information across interactive frames, generating multi-frame masks. Additionally, we leverage an id-queried mechanism to process multiple objects. To achieve more efficient propagation and a lightweight model, we propose a truncated re-propagation strategy, replacing the previous multi-round fusion module, which employs an across-round memory that stores crucial interaction information. Our SwinB-IDPro attains a new state-of-the-art performance on DAVIS 2017 (89.6\%, $\mathcal {J\&F}@60$). Furthermore, our R50-IDPro exhibits over $\bf{3 \times}$ faster performance than the leading competitor in challenging multi-object scenarios.
\end{abstract}


\section{Introduction}
\label{sec:intro}
\IEEEPARstart{I}{nteractive} Video Object Segmentation (iVOS) aims to produce real-time segmentation of objects during user interactions, with numerous downstream applications such as video editing and object tracking. Unlike traditional VOS tasks, iVOS requires real-time human-computer interaction, which means that the user's input habits, system running time, and memory requirements are all important considerations.

Previous methods have made significant progress. The coupling model~\cite{OhLXK19, HeoKK20, MiaoWY20,liang2020memory} jointly trains the interaction and propagation modules, but these coupling models might be trained hard. Then, the decoupled models~\cite{abs-1801-00269, ChengTT21, zhu2021separable} were proposed, which train the interaction and propagation modules separately, such as the recent SOTA method that introduces a difference-aware fusion module~\cite{ChengTT21}.

However, all current methods suffer from two key limitations:  {\bf{i) Single Input Mode.}} They are constrained by a single input mode, whereby only single-frame annotations can be accommodated at a given instance. This limitation hampers the user's capacity to articulate their intentions comprehensively. Recognizing the user's inclination to annotate multiple frames sequentially, as evident in scenarios where they annotate a cat in the third frame and subsequently a dog in the seventh frame while viewing a video, the inefficiency of processing only one frame of interaction per update round becomes apparent. This constraint impedes the seamless annotation of evolving content and warrants a more flexible approach to user input in annotation processes. {\bf{ii) Slow Running Speed.}} Most recent methods process multiple objects independently and then fusion all the single-object predictions into a multi-object prediction, as demonstrated in Fig.~\ref{fig:introduction}. Such individual processing consumes $F \times N$ running time~($F$ denotes the number of frames and $N$ denotes the number of objects), increasing linearly with the number of objects to be segmented. 

To address limitations, we design targeted architecture:

{\bf{i) Flexible Multiple Input Modes}}. Ideally, iVOS system should accommodate both single-frame and multi-frame scribbles to provide users with versatile input options. In multi-frame scenarios, users commonly identify distinct objects across frames, annotating a person in one frame and a horse in another, for instance. To address the segmentation of these varied objects across frames, we introduce the across-frame interaction module~(AFI), which facilitates the seamless transfer of scribble-based information between interactive frames, enabling the simultaneous decoding of masks associated with multiple interactive frames. This module ensures effective processing of annotations across different frames, promoting a more comprehensive and flexible annotation experience.

{\bf{ii) ID-Queried Concurrent Propagation}}. Previous methods suffer from slow running speed as they generate masks for each object individually. To enhance processing speed, a promising approach is to devise a model capable of handling multiple objects simultaneously. Nevertheless, a challenge lies in effectively distinguishing between these objects in a unified manner. To address this issue, we propose to assign a unique identifying tag to each object and train multiple distinct channels to efficiently process the associated scribbles and mask information linked to each tag. This id-queried Concurrent method significantly reduces the running time from $F \times N$ to approximately $F$ when dealing with $N$ objects.

On the whole, we propose a flexible iVOS framework by  \textbf{ID}-queried Concurrent \textbf{Pro}pagation~(IDPro; Fig.\ref{fig:pipeline}), which contains two main components: Across-Frame Interaction Module~(AFI) and Concurrent Propagation Module. The AFI module generates masks for the interacted frames, while the Concurrent Propagation Module handles the remaining non-interactive frames. Concretely, we design the interactive module~(see Fig.\ref{fig:s2m}) that can accept distinct scribbles on multiple frames and migrate scribbles information across frames, thus we can generate masks for all annotated objects and interacted frames based on the input scribbles. Moreover, during propagation, we propose an innovative across-round memory to store the important interacted information and update it for each round to generate masks. In addition, the truncated propagation strategy proposed can fully use across-round information by re-propagating for all frames.

Extensive experiments are conducted to validate the effectiveness. Particularly, our SwinB-IDPro achieves 89.6\% $\mathcal {J\&F}@60$ on DAVIS after only three rounds of interaction. Meanwhile, our R50-IDPro runs more than {\bf $\bf{3 \times}$ faster} compared to the state-of-the-art method under challenging multi-object scenarios. Our code and benchmark will be released.

Overall, our contributions are summarized as follows:
\begin{itemize}
\vspace{-0.1cm}
\setlength{\itemsep}{0pt}
\setlength{\parsep}{0pt}
\setlength{\parskip}{0pt}

\item We are the first to propose a framework called IDPro that enables users to annotate distinct objects on multiple frames in a single round, broadening the application scenarios and aligning better with people's usage habits. Additionally, we introduced an interactive graphical user interface~(GUI) that supports multi-frame interaction.

\item Our proposed Across-Frame Interaction Module is a novel approach that facilitates the transfer of scribble-based information between interactive frames. Moreover, we develop an Concurrent Propagation Module that can simultaneously process multiple objects, and design a re-propagation strategy to effectively exploit the interactive information from multiple rounds.

\item We conduct extensive experiments on DAVIS 2017~\cite{Pont-TusetPCASG17} and the experimental results strongly demonstrate that our proposed IDPro runs {\bf $\bf{3 \times}$ faster} faster than the state-of-the-art method under challenging multi-object scenarios.

\end{itemize}

\section{Related Work} 
\subsection{Video Object Segmentation}
The objective of Video Object Segmentation~\cite{cheng2021rethinking, zhuangwei2021survey, yao2020video, cheng2023segment, chen2017deeplab, ge2021video, wang2021ainet, yin2017joint, grundmann2010efficient, faktor2014video, brox2010object, zou2024segment, cheng2022xmem, yang2022decoupling,zhou2020matnet, zhou2020motion, liang2023local} is to derive masks for target objects across the entire video. Methodologies commonly fall into the categories of semi-supervised VOS and unsupervised VOS. Unsupervised Video Object Segmentation (UVOS) specifically aims to autonomously distinguish primary foreground object(s) from the background in a video sequence. Existing methodologies have undergone extensive exploration. Notably, IMC-Net~\cite{xi2022implicit} introduces an implicit motion-compensated network that adeptly integrates complementary appearance and motion cues. In addition, \cite{li2023adversarial} puts forth a contrastive motion clustering algorithm, while FEM-Net~\cite{zhou2021flow} introduces a motion-attentive encoder and a Flow Edge Connect module to infer edges in the optical flow for ambiguous or missing regions.

Another prevalent approach is semi-supervised VOS~\cite{van2020survey}, semi-supervised video object segmentation involves segmenting a specific object with a fully-annotated mask in the initial frame. Early methods~\cite{zhou2022survey, ChengTT21, VoigtlaenderL17, KhorevaBIBS19, PerazziKBSS17, DBLP:journals/corr/abs-1812-01233, DBLP:journals/corr/abs-1806-06157, DBLP:journals/corr/abs-2103-15662, DBLP:journals/corr/abs-1807-10982, DBLP:journals/corr/abs-2110-06915, xu2022switch} predominantly focus on modeling appearance representations and temporal connections. MoNet~\cite{XiaoFLLZ18} fine-tunes pre-trained networks based on first-frame ground truth during testing, while OnAVOS~\cite{voigtlaender2017online} extends the first-frame fine-tuning by introducing an online adaptation mechanism. Similarly, MaskTrack~\cite{PerazziKBSS17} and PReM~\cite{luiten2018premvos} utilize optical flow to propagate segmentation masks across frames. Nonetheless, these fine-tuning methods exhibit inefficiencies.

Recent advancements in video object segmentation methods have showcased innovative approaches~\cite{lan2022siamese, Vujasinovic_2022, yin2021learning, HeoKK20, MiaoWY20, cheng2021rethinking}. For instance, \cite{zhu2021separable} proposes an approach that considers pixel-wise similarities between reference and target frames, alongside the structural information of target objects. Similarly, OGS~\cite{fan2021semi} introduces an innovative architecture based on object-aware global-local correspondence. PML~\cite{ChenPMG18} utilizes the nearest neighbor classifier to learn pixel-wise embedding, while VideoMatch~\cite{HuHS18a} employs a soft matching layer, mapping current frame pixels to those in the initial frame. Building upon these foundations, CFBI~\cite{abs-2010-06349, YangWY22} extend the pixel-level matching mechanism by additionally matching between the current frame and the previous frame. STM~\cite{OhLXK19} adopts a memory bank constructed from past frames and integrates a query key-value attention mechanism. Notably, AOT~\cite{yang2021associating} introduces an identification mechanism to enable the simultaneous processing of multiple objects within a frame.

To address multiple object segmentation in videos, traditional interactive methods process each object separately with a decoder before integrating the segmentation outcomes. However, these methods are inefficient, impacting user interactions negatively. To align better with people’s usage habits, we present an interactive segmentation technique that can simultaneously process multiple objects. This approach notably boosts processing speed, consequently elevating the overall smoothness of the interactive system experience.

\subsection{Interactive Video Object Segmentation (iVOS)}
\lkx{In contrast to other cross-modal video segmentation tasks like text-guided~\cite{gao2023decoupling, 10146303, gao2021clip, wu2022language, botach2022end, zhang2021tipadapter, kim2021vilt, li2023refsam} or audio-guided video segmentation~\cite{li2023catr, huang2023discovering, hao2024improving, chen2024unraveling, yang2023cooperation}, interactive video object segmentation~(iVOS) methods~\cite{heo2021guided, xu2016deep, varga2021fast, Chen_Zhao_Zhang_Duan_Qi_Zhao_2022, Liu_2023_ICCV} afford greater flexibility in incorporating user guidance information.} Interactive video object segmentation aims to produce real-time segmentation of a target object during user interactions, the interactions can be used to either segment an object or correct previously misclassified region. 

Early methods use hand-crafted features (e.g.,click-based interactions~\cite{WangHC14,JainG16}), but they require a large number of user interactions, making them impractical at a large scale~\cite{Garcia-GarciaOO18}. With the release of the DAVIS interactive benchmark~\cite{abs-1803-00557, Caelles_Montes_Maninis_Chen_Gool_Perazzi_Pont-Tuset_2018}, a standardization for the evaluation of iVOS has been introduced. Previous approaches in user-guided segmentation often employed deep feature fusion techniques, utilizing interconnected encoder networks~\cite{sun2016interactive, oh2019fast, heo2019interactive, HeoKK20}, or storing scribble features in memory for subsequent reference during segmentation~\cite{oh2020space, liang2020memory, MiaoWY20}. However, these methods inherently link the specific user inputs with the mask propagation process~\cite{oh2019fast}, complicating model training. Balancing the understanding of user interactions and accurate mask propagation simultaneously poses challenges. 

\lkx{To optimize training procedures and improve segmentation precision, recent approaches have drawn inspiration from object tracking methodologies~\cite{xu2022switch, wang2020towards, luo2021multiple, yao2020video, cheng2023segment, yin2017joint, brendel2009video, rajivc2023segment}. These iVOS methods endeavor to disentangle interaction and propagation networks, as delineated in prior works~\cite{benard2017interactive, tran2020interactive}, which entails the creation of a mask through diverse interaction modalities, subsequently propagated for refined segmentation outcomes.} For instance, IPN~\cite{OhLXK19} and ATNet~\cite{HeoKK20} use two segmentation networks to handle interaction and propagation, respectively. Cheng et al.~\cite{ChengTT21} explore the differences in mask domains pre and post-interaction rounds, further refining the decoupling model.

While current decoupling methods boast higher accuracy, they fall short in enabling simultaneous user interaction across multiple frames within a decoupled framework. This limitation restricts both model efficiency and the expression of user intent. Thus, there is a pressing need for a framework capable of processing one or more scribbles concurrently.

\begin{figure*}[t]
    \centering
    \includegraphics[width=1\textwidth]{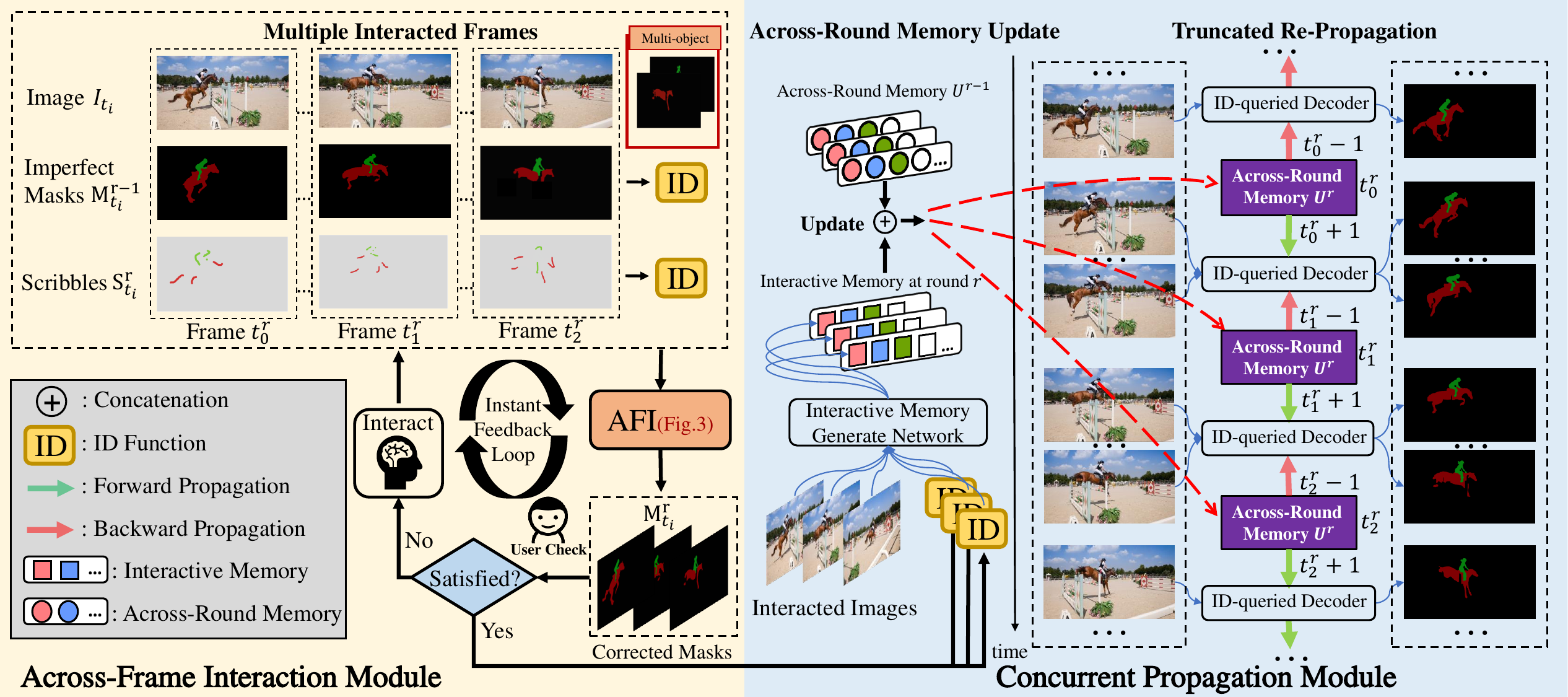}
    \vspace{-0.5cm}
    \caption {{\bf The overview of IDPro~(\S\ref{sec:overview})}. IDPro can be formulated as interaction-propagation. During interaction round $r$, the Across-Frame Interaction Module enables users to annotate multiple frames and generate the corresponding masks. Next, the Concurrent Propagation Module is designed to generate the masks for the non-interactive frames. To preserve crucial interaction information, we update the across-round memory after each round. Then we generate masks with the updated memory by the ID-queried decoder. We further introduce a re-propagation strategy to address conflicts in multi-round propagation.}
    \vspace{-0.5cm}
    \label{fig:pipeline} 
\end{figure*}

\section{Approach}

\subsection{Overview}
\label{sec:overview}
Our pipeline for iVOS task can be formulated as interaction-propagation~(depicted in Fig.~\ref{fig:pipeline}). To address limitations in previous methods, such as single input mode and slow running speed, we carefully design two modules: the Across-Frame Interaction Module~(AFI) and the Concurrent Propagation Module. These modules enable effective interactions across frames and facilitate information propagation across rounds.
~\\

\noindent {\bf{Across-Frame Interaction Module~(AFI).}}
The Across-Frame Interaction Module (AFI; detailed in Sec.\ref{sec:s2m}) stands out from current methodologies reliant on single-frame inputs. In contrast, our AFI module excels in the segmentation of multiple frames and diverse objects, showcasing its capacity to handle concurrent propagation and enhance efficiency.

We use $r$ to denote the current interaction round, ${\bf R}=\{t_i|i=0,1,…,T\}$ to denote the set of interacted frames, and $\widetilde{\bf R}$ to denote the set of frames without interaction.

The goal of the AFI module is to generate the masks $M_{t}^r,{t} \in {\bf R}$ for the interacted frames at round $r$.
\begin{equation}
\{M_{t_i}^r\}_{i=1}^T = AFI(\{{S_{t_i}^r}, M_{{t_i}}^{r-1},  {I_{t_i}}\}_{i=1}^T), {t_i} \in {\bf R},
\end{equation}
where $AFI(\cdot)$ is the AFI module, which is utilized with scribbles, previous masks, and RGB images of multiple interacted frames, denoted as $\{{S_{t_i}^r}, M_{{t_i}}^{r-1},  {I_{t_i}}\}_{i=1}^T$, to generate output masks, denoted as $\{M_{t_i}^r\}_{i=1}^T$. If it is the first round of interaction, the previous masks are empty. Conversely, if it is a round following an interaction, the previous masks contain problems that require correction. \lkx{This mask, derived from the preceding interaction and requiring refinement, is referred to as the imperfect mask.}
~\\

\noindent {\bf{Concurrent Propagation Module}  } 
The Concurrent Propagation Module aims to generate masks for multiple non-interactive frames. In contrast to the previous propagation module that relied solely on the current round information and processes objects individually, our module leverages cross-round information and handles multiple objects in parallel. Concretely, we propose the Concurrent Propagation Module $Propagate(\cdot)$ to produce masks,
\begin{equation}
M_{t_j}^r = Propagate(M_{t_i}^r, I), {t_i} \in {\bf R}, {t_j} \in \widetilde{\bf R},
\end{equation}
where $ M_{t_i}^r $ is the output mask of  $AFI(\cdot)$ at $t_i$-th frame and $r$-th round, $ M_{t_j}^r $ is the mask result at $t_j$-th non-interactive frame and $r$-th round, and $I$ denotes the input images.

\begin{figure*}[t]
    \centering
    \includegraphics[width=0.95\textwidth]{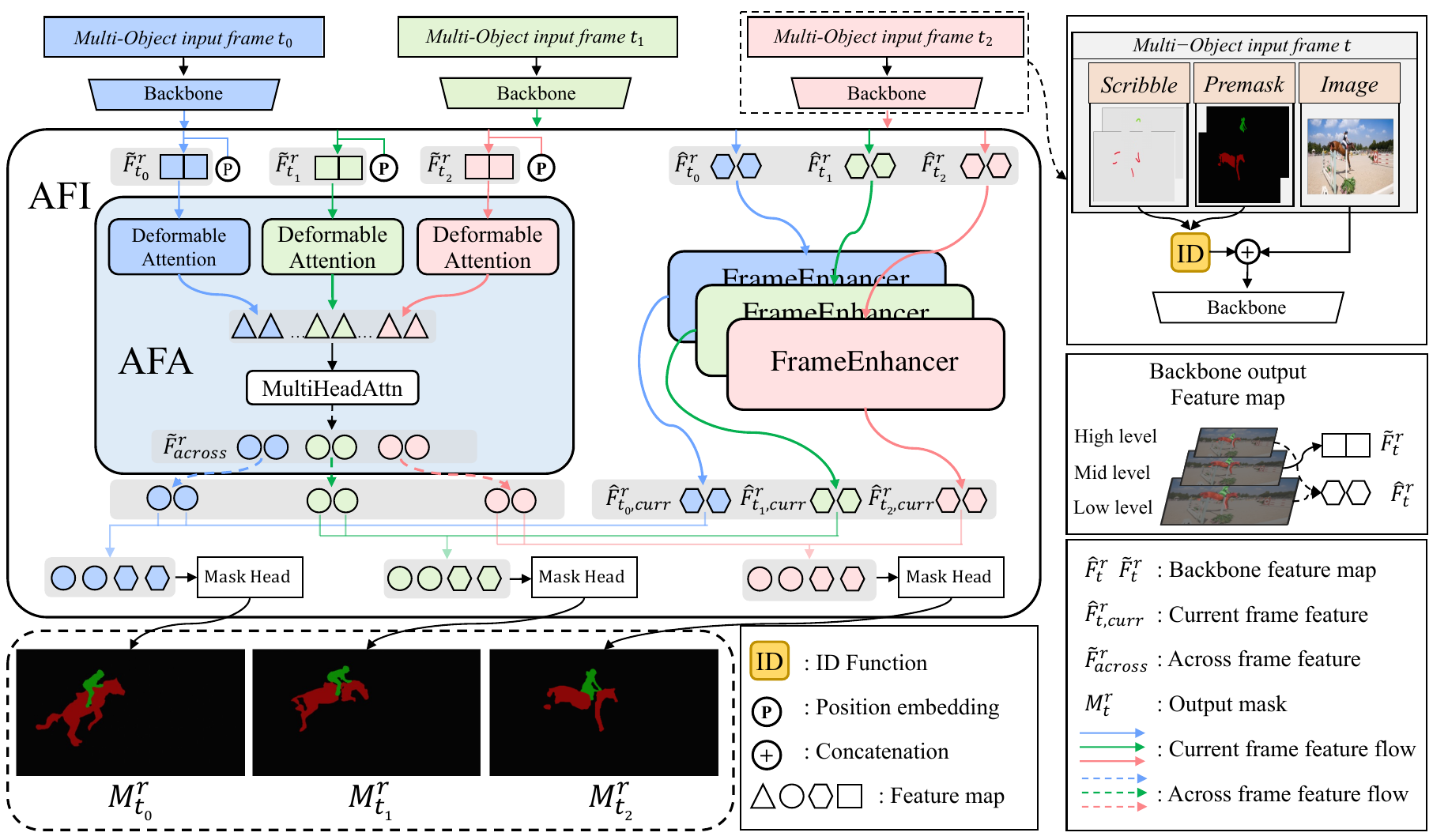}
    \vspace{-0.3cm}
    \caption{{\bf The implementation of Across-Frame Interaction Module~(AFI; \S\ref{sec:s2m})}. To capture the multi-scale features within each frame and the temporal information across frames, we design an across-frame attention~(AFA). To enhance the modeling of the current frame, we develop the FrameEnhancer Network. The premask at the top right of the image denotes the previous mask.}
    \vspace{-0.2cm}
    \label{fig:s2m}
\end{figure*}

\begin{figure}[t]
    \centering
    \includegraphics[width=0.48\textwidth]{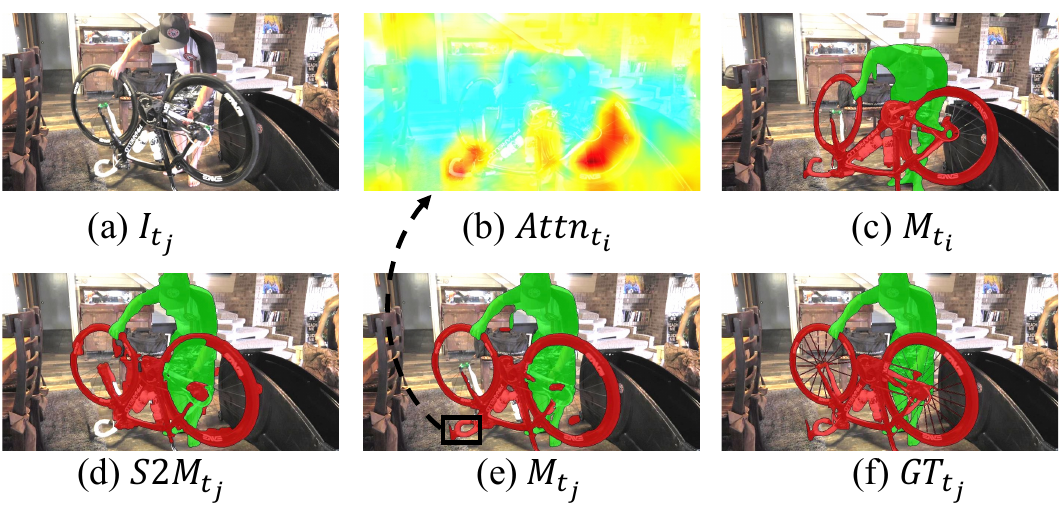}
    \vspace{-0.3cm}
    \caption{{\bf The attention map of Across-Frame Attention~(AFA; \S\ref{sec:afa})}. Continuing Fig.~\ref{fig:s2m}, showing the S2M in MiVOS that accepts only one single frame is insufficient. Suppose we aim to predict the mask of $t_{j}$-th frame. (d) The mask produced by S2M only refers to the $t_{j}$-th frame loss of the handlebar. To address this challenge, we utilize across-frame features to generate an attention map~(b), which represents the handlebar's attention position in the $t_{j}$-th frame~(e) on the $t_{i}$-th frame~(c). Thus, the mask~(e) produced by IDPro contains the information of the other interacted frame~({e.g. the handlebar information at $t_{i}$-th frame}).}
    \vspace{-0.3cm}
    \label{fig:compare}
\end{figure}

\subsection{Across-Frame Interaction Module~(AFI)}
\label{sec:s2m}
In contrast to existing methods that are restricted to single-frame and single-object scenarios, we propose the Across-Frame Interaction Module~(AFI; Fig.~\ref{fig:s2m}), which is a novel method that can handle multiple frames and objects simultaneously. The AFI module comprises two sub-components: the Across-Frame Attention sub-component~(AFA) and the FrameEnhancer sub-component. The AFA sub-component captures temporal dependencies among frames, while the FrameEnhancer sub-component extracts salient features from each frame.
~\\

\noindent {\bf{Identification Mechanism and Encoder.}}
To achieve simultaneous processing of multiple objects, we employ an identification mechanism on both masks and scribble maps. This enables us to obtain a mask, denoted as $M_{{t}}^{r} \in \mathbb{R}^{T H W \times C}$, and a scribbled map, denoted as $S_{{t}}^{r} \in \mathbb{R}^{T H W \times C}$, both with a unique identification tag. For each interacted frame, we concatenate the scribble $S_{{t}}^{r}$, mask $M_{{t}}^{r-1}$, and RGB image ${I_{t}}$, and feed them into the backbone $Encoder(\cdot)$. Then we obtain the multi-scale feature ${F}_{t}^{r}$.
\begin{equation}
{F}_{t}^{r} = Encoder(Concat({S_{{t}}^{r}}, {M_{{t}}^{r-1}},{I_{t}})), {t} \in {\bf R},
\label{eq:encoder}
\end{equation}
where $Concat(\cdot)$ denotes the concatenate operation.

After obtaining the multi-scale features ${F}_{t}^{r}$, we divide them into two parts: ${\{\widetilde{F}_{t}^{r}, {\hat{F}}_{t}^{r}}\}$. The middle-level feature, $\widetilde{F}_{t}^{r}$, captures across-frame information in the AFA sub-component, while the high-level and low-level feature, ${\hat{F}_{t}^{r}}$, extracts current-frame information in the FrameEnhancer sub-component. This enables us to incorporate both the inter-frame and intra-frame information in our model.
~\\

\noindent {\bf{Migrate Across-Frame Scribbles Information.}} 
\label{sec:afa}
To capture temporal information from multi-frame scribbles, we design a novel across-frame attention~(AFA) that integrates a deformable attention mechanism~\cite{zhu2020deformable} and a multi-head attention mechanism~\cite{vaswani2017attention}. \lkx{The AFA module is tasked with the simultaneous processing of information from multiple frames, necessitating considerable computational resources. In order to reduce the computational load while retaining the texture information of the image, we have chosen to use mid-level image features. Furthermore, since mid-level feature maps contain less information compared to low-level feature maps, and the attention mechanism can effectively capture global information in the image, we designed to use the attention mechanism to aggregate features over a larger range, thereby capturing the global relationships in the image.} The deformable attention allows for information exchange across different scales within each frame, while the multi-head attention facilitates information exchange between frames. This approach enhances the temporal coherence of the scribbles and enables effective modeling of objects.
\begin{equation}
\begin{aligned}
\begin{split}
Q = K &= V = \{DA(\widetilde{F}_{t_i}^{r})\}_{i=0}^T,  {t_i}\in {\bf R}, \\
{\widetilde{F}}_{across}^{r}&=MultiHead(Q,K,V),
\end{split}
\end{aligned}
\end{equation}
where $DA(\cdot)$ is deformable attention, which is employed as self-attention within each frame. $MultiHead(\cdot)$ is multi-head attention, employed as cross-attention between different frames. The Q, K, and V embeddings, after undergoing deformable attention, aggregate multi-scale semantic information for frames respectively, and the across-frame feature embedding $\widetilde{F}_{across}^{r}$ captures information across interacted frames.

To demonstrate the effectiveness of the across-frame attention, we visualized the attention map~(see Fig.~\ref{fig:compare}).
~\\

\noindent {\bf{Multi-Frame Masks Generation.}} 
To improve feature utilization and model the current interactive frame more effectively, we propose a FrameEnhancer sub-component. \lkx{Concretely, low-level features contain a wealth of edge and texture information, while high-level features are rich in semantic information. Thus, we combine the low-level and the high-level features to better model the information of the current frame. Furthermore, convolution operations are efficient in processing high-resolution feature maps and can extract low-level features such as edges and textures from images through local receptive fields, they are more conducive to modeling the current frame. Therefore, we have applied convolution operations in the design of the FrameEnhancer sub-component.}
\begin{equation}
{\hat{F}_{t, curr}^{r}} = \mathcal{G}({\hat{F}_{t}^{r}}), {t} \in {\bf R},
\end{equation}
where $\mathcal{G}(\cdot)$ denotes the FrameEnhancer sub-component, which includes ASPP and FFN. ${\hat{F}_{t, curr}^{r}}$ is the feature embedding of current $t$ frame.

The across-frame feature embedding ${\widetilde{F}}_{across}^{r}$ and the current-frame feature embedding ${\hat{F}}_{t, curr}^{r}$ will be used to predict the corresponding masks.

We implement a simple network as our mask head, incorporating Conv2D, BatchNorm, and Relu. The final mask $M_{t}^r, {t} \in {R}$ is predicted using both the across-frame feature embedding ${\widetilde{F}}_{across}^{r}$ and the current frame feature embedding ${\hat{F}}_{t, curr}^{r}$ with the mask head.
\begin{equation}
M_{t}^r = MaskHead( {\widetilde{F}}_{across}^{r}, {\hat{F}}_{t, curr}^{r}),{t} \in {\bf R}.
\end{equation}

For more details of the AFI module, refer to Fig.~\ref{fig:s2m}. 

\subsection{Concurrent Propagation Module}
\label{sec:prop}

Previous methods have primarily focused on modeling the current round while ignoring the interaction between different rounds. In light of this, we design the Concurrent Propagation Module, which maintains an across-round memory and exploits it for propagation. Moreover, we designed an ID-queried decoder that can simultaneously process multi-object masks and images with identification tags in an efficient manner. The Concurrent Propagation Module is designed to generate masks for the non-interaction frames. To begin, we extract information from the interacted frame of the current round to obtain the interactive memory and employ it to update the across-round memory. Then we produce masks with the ID-queried decoder by employing the across-round memory. In addition, we propose a truncated propagation strategy to address conflicts and an innovative re-propagation strategy to improve segmentation quality.
~\\

\noindent {\bf Across-Round Memory Update.} 
We contend that information from interacted frames is particularly trustworthy, as it reflects the user's intentions captured by the interaction module. Thus, we designed the across-round memory to store the interacted frame informations in $0\sim r$ rounds, which serves as a reference for subsequent rounds. The across-round memory, denoted as ${ U}^{r}$, is updated iteratively.
\begin{equation}
{ U}^r = {Update}(\mathcal{A}(I_{t}, M_{{t}}^{r}), { U}^{r-1}), {t} \in {\bf R},
\end{equation}
where $\mathcal{A}(\cdot)$ denotes the Concurrent Memory Update network~(MU-Net, adopted AOT). ${ U}^{r}$ denotes the across-round memory in the $r$-th interaction round, $I_{t} \in \mathbb{R}^{T H W \times C}$ is the RGB image of the interacted frame, and $M_{{t}}^{r} \in \mathbb{R}^{T H W \times C}$ is the mask of the interacted frames with identification tags.
~\\

\noindent {\bf Propagation with ID-queried Decoder.}
We put the image and the across-frame memory into the ID-queried decoder on a frame-by-frame basis to generate the corresponding multi-object mask. The ID-queried decoder contains multiple object-level channels that can simultaneously process multiple objects with identification tags,
\begin{equation}
M_{t}^r = \mathcal{P}({\bf U}^r, I_{t}), {t} \in {\bf \widetilde R},
\end{equation}
where $\mathcal{P}(\cdot) $ denotes the ID-queried decoder. $I_{t} \in \mathbb{R}^{T H W \times C}$ denotes the RGB image of $t$-th frame that without interaction. ${\bf U}^r$ denotes the across-frame memory and $M_{t}^r \in \mathbb{R}^{T H W \times C}$ denotes the output mask of $t$-th non-interactive frame. $\widetilde{\bf R}$ is the set of frames without interaction.
~\\

\noindent {\bf Truncated Strategy.}
In the previous study~\cite{ChengTT21}, where only one interacted frame existed, employing the bidirectional propagation didn't encounter conflicts in a single round. However, in our setting, there are multiple interacted frames, and employing bidirectional propagation inevitably leads to conflicts both within a single round and between different rounds.

\begin{figure}[t]
    \centering
    \includegraphics[width=0.48\textwidth]{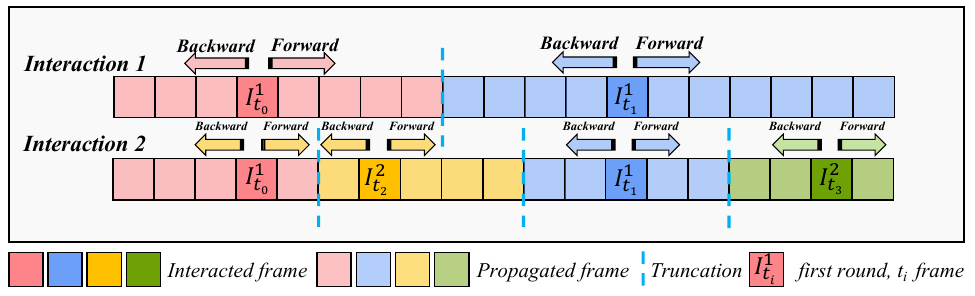}
    \vspace{-0.2cm}
    \caption{{\bf The illustration of Truncated Propagation and Re-propagation Strategy~(\S\ref{sec:prop})}. To overcome the conflicts, we propose the truncated propagation that propagates the masks of the interacted frames to the midpoint of the two frames. For re-propagation, all frames without interaction will be re-propagated with the more robust across-round memory that we designed.}
    \vspace{-0.5cm}
    \label{fig:truncate}
\end{figure}

To address these conflicts, MiVOS designed a fusion module to merge two masks, but we think it is insufficient. Thus, we designed a simple but effective truncation propagation strategy that replaces the fusion module. Assuming that the user interacts with multiple frames, the propagation network propagates masks starting from these interacted frames and stops in the middle of two interacted frames~(refer to Fig.~\ref{fig:truncate}).
~\\

\noindent {\bf Re-propagation.} 
The across-round memory ${ U}^{r}$ incorporates all user interactions from $0\sim r$ rounds, which is more robust than the previous memory ${ U}^{r-1}$ as it contains information from the $r$-th round. However, existing methods only propagate a subset of frames in the current round, leaving some frames unpropagated. Thus, we design to re-propagate all non-interacted frames with the improved and more reliable ${ U}^{r}$, resulting in enhanced performance.

\section{Implementation Details}

\noindent {\bf Datasets.}
Our model is trained on the  Lvis~\cite{gupta2019lvis}~(static image dataset), YouTubeVOS~(YTB)~\cite{xu2018youtube} and DAVIS 2017~\cite{pont20172017}. The three datasets vary in quantity and quality of annotations: 1) {\bf Lvis} is designed for large vocabulary instance segmentation, containing instance masks for over 1000 entry-level object categories. We obtained multiple images from the static images in Lvis by applying translations, rotations, and other morphological changes. We then treated these images as a video and conducted pre-training for IDPro using this modified dataset. 2) {\bf YouTube-VOS} is the latest large-scale benchmark for multi-object video segmentation. Specifically, YouTube-VOS contains 3471 videos in the training split with 65 categories and 474/507 videos in the validation 2018/2019 split with an additional 26 unseen categories. The unseen categories do not exist in the training split to evaluate the algorithms’ generalization ability. 3) {\bf DAVIS 2017} contains 60 training sequences and 30 validate sequences. We use the DAVIS 2017 training set, which provides high-quality densely annotated segmentation mask annotation for each frame. For evaluation, we employ 30 validate sequences.

\begin{figure}[t]
    \centering
    \includegraphics[width=0.48\textwidth]{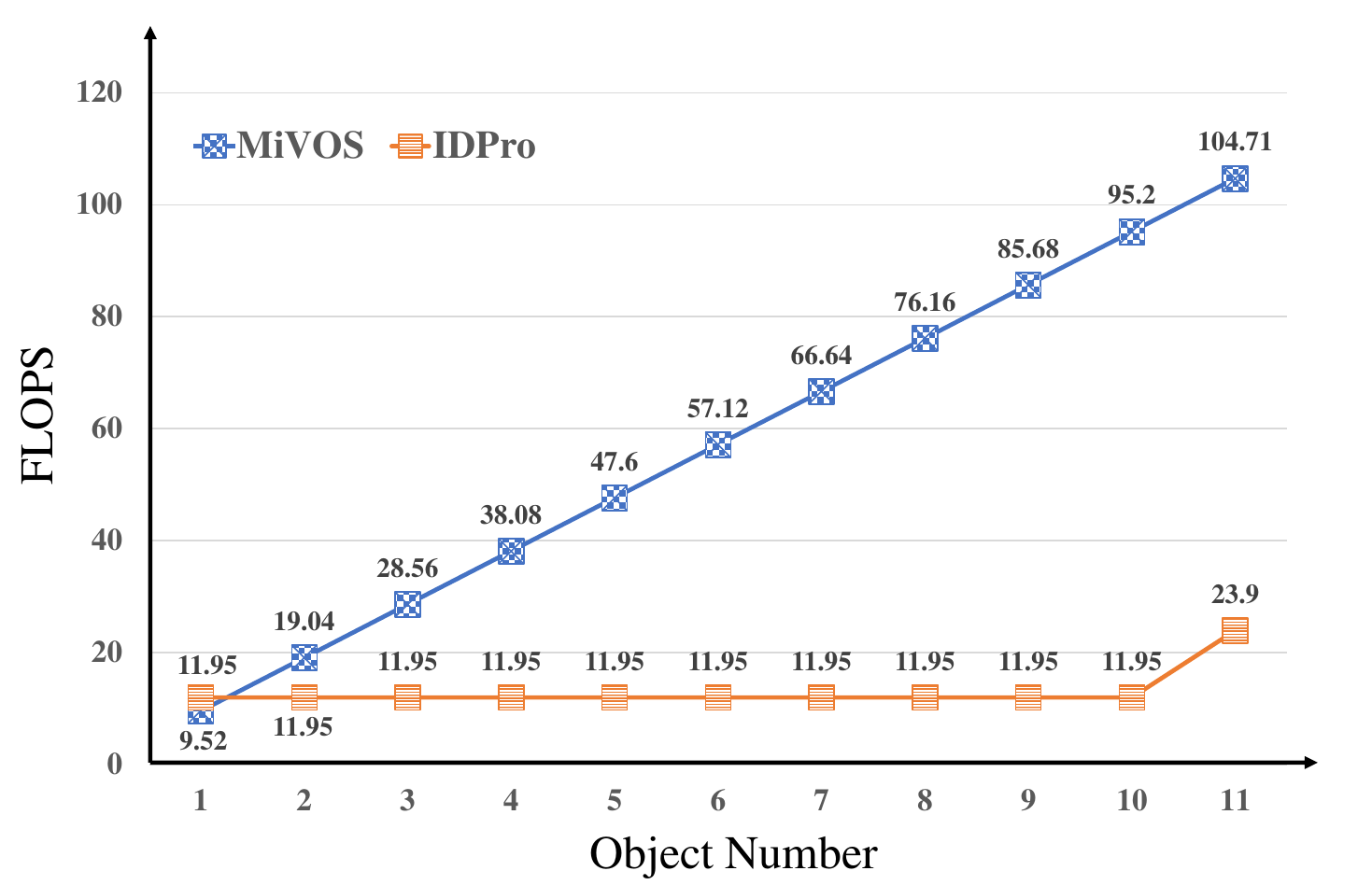}
    \vspace{-0.2cm}
    \caption{{\bf The computational complexity varies with the number of objects~(\S\ref{sec:speed})}. \lkx{``FLOPS'' stands for ``Floating Point Operations Per Second'' and is a measure of the computational complexity of a system. The computational complexity of MiVOS is increased with the number of objects, but IDPro processes multiple objects as efficiently as a single object. Note that we set the batch in the decoder as 10, so the time consumed increases as the object number is greater than 10, and we have the flexibility to train models with different batch numbers as needed.}}
    \label{fig:zhe}
    \vspace{-0.4cm}
\end{figure}

\begin{figure}[t]
    \centering
    \includegraphics[width=0.48\textwidth]{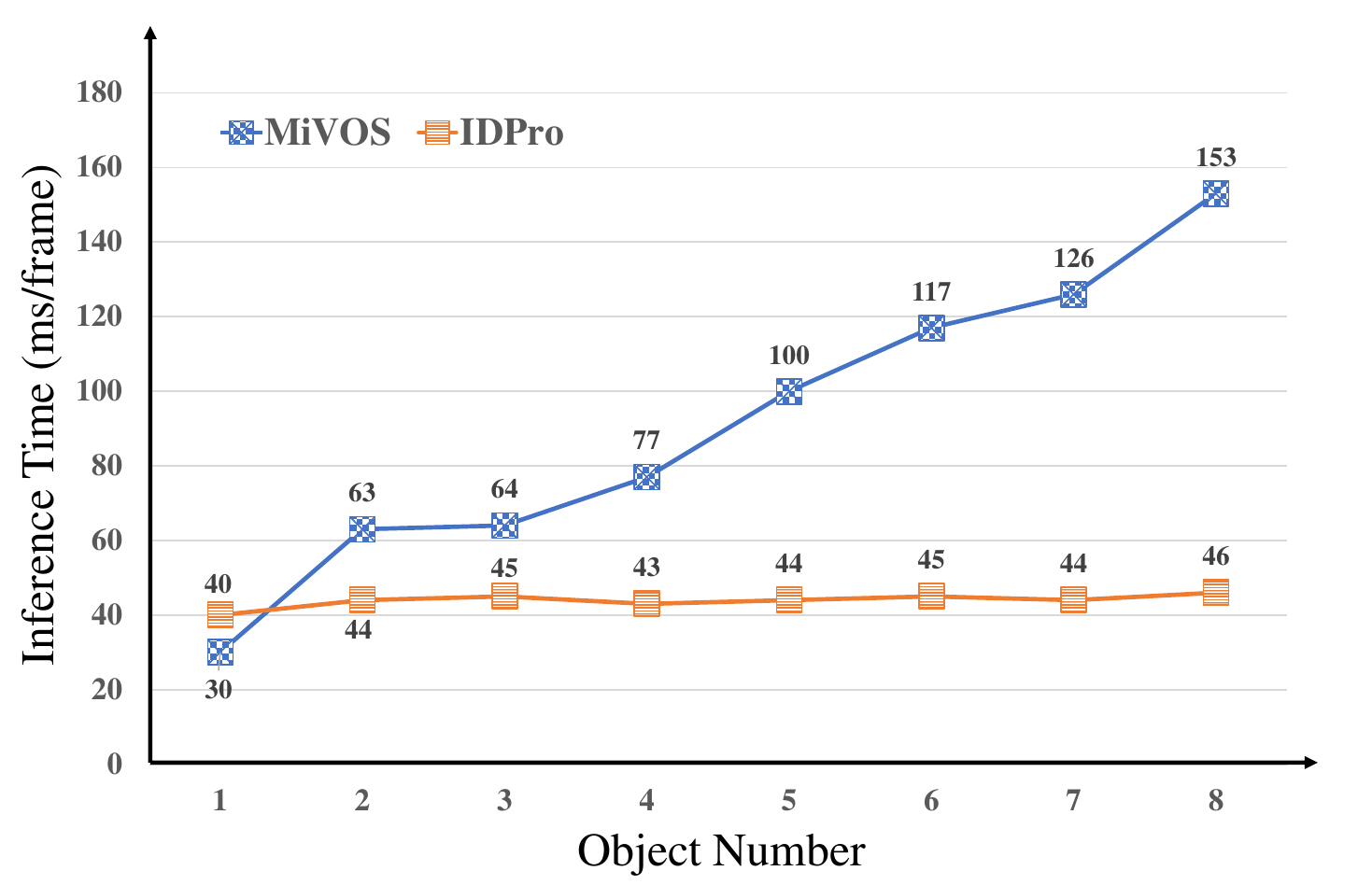}
    \vspace{-0.2cm}
    \caption{{\bf The inference speed varies with the number of objects~(\S\ref{sec:speed})}. \lkx{The time consumption of MiVOS is increased with the number of objects, but IDPro processes multiple objects as efficiently as a single object. }}
    \label{fig:zhe}
    \vspace{-0.4cm}
\end{figure}

To evaluate the effectiveness of IDPro, we conducted experiments on the DAVIS 2017. Currently, DAVIS is the sole benchmark for systematically validating iVOS. It offers diverse scribble types, automatic scribble generation, and an interactive evaluation environment. Thus, we follow previous SOTA work to develop IDPro on it.
\\~

\noindent {\bf Scribbles Generation.} \lkx{The robots in DAVIS can provide scribbles for a selected frame. We generated scribbles
for YouTube-VOS~(YTB) consistent with the DAVIS methodology. To ensure fairness during inference, we adopt the DAVIS test setting. Specifically, we use three initial scribbled frames provided by DAVIS, and use the toolkit to select the worst predicted frame and scribble it.}
\\~

\noindent {\bf Evaluation Metric.}
We use $\mathcal {J\&F}@60$ as the evaluation metric, where $\mathcal {J\&F}$ refers to Jaccard and F-Score, and $@60$ indicates that the running time of each video must not exceed 60 seconds. In addition, we employ $f$ to denote the number of frames that interacted in one round and use $r$ to denote the number of interaction rounds.
\\~

\noindent {\bf Training Details.}
We implement IDPro by PyTorch, and train all models on a 24GB GPU device~(RTX 3090Ti). When training on the YouTube-VOS and DAVIS datasets, we employ the AdamW optimizer~\cite{kingma2014adam} with a learning rate of $2 \times 10^{-4}$ and $2 \times 10^{-5}$, respectively. Our batch size is set to 4. The AFI Module is first trained on synthetic video sequences by Lvis, then transferred to YouTube-VOS and DAVIS 2017. The loss function is a 0.5:0.5 combination of bootstrapped cross-entropy loss and soft Jaccard loss~\cite{Nowozin14}.

\section{Experiments}
\subsection{Comparison at Speed.}
\label{sec:speed}
\noindent\textbf{Settings.} In Fig.~\ref{fig:zhe}, we compare the inference speed of IDPro with the state-of-the-art model MiVOS, across varying numbers of target objects. The maximum number of objects IDPro can concurrently handle is dictated by the channel quantity established during training, set in this instance to accommodate up to ten objects simultaneously. Consequently, we present an experiment of inference speed, from processing a single object to concurrently handling eleven objects, clearly illustrating the model's inference speed modulation in response to increasing object quantity.
\\~

\begin{table}[]
\centering
\caption{{\bf The performance on the DAVIS 2017~(\S\ref{sec:acc})}. The $f$ denotes the number of frames interacted in one round, and $r$ denotes the number of interaction rounds. (+Y) denotes model pre-trained IDPro on YTB. ${\dagger}$ denotes that we trained IDPro under a single-frame scenario. The best results are shown in \textcolor{red}{red}.}

\begin{tabular}{l|c|c}
\toprule
\multicolumn{3}{c}{\bf{Old Setting: Single-Frame ($f=1$, $\mathcal {J\&F}@60$)}}                        \\  \midrule
Model             & $r=1$ & $r=3$                         \\  \hline
\lkx{SAM-PT}~\cite{rajivc2023segment} & 79.4 & - \\
ATNet~\cite{HeoKK20} & 81.6                           & 82.7                        \\
STM~\cite{OhLXK19} & 82.3                          & 84.8                        \\
\lkx{GNN-anno}~\cite{varga2021fast} & - & 79.0 \\
\lkx{GIS}~\cite{heo2021guided} & 83.6 & 86.6 \\
MiVOS~\cite{ChengTT21}& 85.5                       & {88.5}                        \\ \hline
{\raggedright R50-IDPro}    & 85.2                        & 87.4                        \\
{\raggedright SwinB-IDPro}  & 86.2                        & 88.2                        \\
R50-IDPro (+Y)   & 85.4                        & 87.3                        \\
SwinB-IDPro$^{\dagger}$ (+Y) & {\textcolor{red}{86.2}} & {\textcolor{red}{88.7}} \\ \midrule \midrule
\multicolumn{3}{c}{\bf{New Setting: Multi-Frame ($f=3$, $\mathcal {J\&F}@60$)}}                         \\ \midrule
Model             & $r=1$                         & $r=3$                         \\   \hline
R50-IDPro          & 86.9                        & 88.6                        \\
SwinB-IDPro        & 87.5                        & 89.3                        \\
R50-IDPro (+Y)   & 87.0                         & 89.1                        \\
SwinB-IDPro (+Y) & {\textcolor{red}{87.5}} & {\textcolor{red}{89.6}} \\ \bottomrule
\end{tabular}
\vspace{-0.1cm}
\label{tab:all}
\end{table}

\begin{table}[t]
\centering
\caption{{\bf The ablation study on the DAVIS with multi-frame and multi-round scenario~($f=3$, $r=3$)~(\S\ref{sec:ablation})}. We first trained IDPro on Lvis and we took a network that without Across-Frame Attention and Re-propagation strategy as the baseline.}
\begin{tabular}{cc}
\toprule
SwinB-IDPro & $\mathcal {J\&F}@60$ \\ \midrule
baseline   & 88.7 \\
(+) Across-Frame Attention        & 89.1 \\
(+) YTB \& DAVIS        & 89.3 \\
(+) Re-propagation Strategy       & 89.6 \\ \bottomrule
\end{tabular}
\label{tab:ablation}
\vspace{-0.4cm}
\end{table}

\noindent\textbf{Results.}
IDPro has an excellent performance in terms of speed. From the results in Fig.~\ref{fig:zhe}, we get the following findings: 1) MiVOS exhibits a progressive rise in inference time correlating with the increase in target objects~(from one object at 9.25 FLOP to ten objects at 95.2 FLOPS). In contrast, our IDPro consistently maintains a steady inference speed of 11.95 FLOPS even in scenarios involving ten objects. Notably, this implies that IDPro can run \textbf{more than 3× faster} than the current state-of-the-art model under challenging multi-object multi-frame scenarios. 2) A notable uptick in inference time for our IDPro model becomes evident when processing the eleventh object. This phenomenon can be attributed to the design of our trained decoder, which operates optimally with a specific batch number of 10, enabling simultaneous processing of up to 10 objects. Consequently, when handling 11 objects, there is an inevitable increase in processing time. To cater to diverse needs, we maintain flexibility in training models with varying batch numbers as required.

\subsection{Comparison at Accuracy.}
\label{sec:acc}
\noindent\textbf{Settings.} 
IDPro can process multi-frame scribble input, but previous methods could only process single-frame scribble input. To ensure fairness, we compared IDPro with competitors under the single-frame setting~($f=1$). Additionally, to validate the effectiveness of our multi-object multi-frame interaction, we also provide the performance of IDPro under the new setting of multiple frames~($f=3$), see Tab.~\ref{tab:all}. 

IDPro introduces a novel capability: processing multi-frame interactive information within a single iteration, a feature absent in previous methods for single iterations. To evaluate this, we conducted experiments, comparing our model against established approaches and the current state-of-the-art method. The assessment involved a dual-setting analysis: 1) \textbf{Old Setting:} Robots could annotate only one frame per iteration and propagation. Following annotation, they received the segmentation result after the initial propagation round. Subsequently, they adjusted annotations for a specific frame, proceeding to the second iteration and propagation, culminating in the segmentation result after the second round. Tab.~\ref{tab:all} presents segmentation results for the first and third rounds in the single-frame scenario. To maintain parity, our IDPro model underwent training in the single-frame datasets. 2) \textbf{New Setting:} Robots could annotate multiple frames per iteration and propagation. After annotating multiple frames, they received the segmentation result after the initial propagation. Subsequently, robots selected multiple frames from the segmentation result for interaction and modification, yielding the segmentation result after the second propagation round. We showcased the segmentation results for the first and third rounds under this new setting, where interactions involved three frames per iteration. The term 'robots' here signifies annotations sourced from the DAVIS 2017 dataset, capable of providing annotations for a selected frame.
\\~

\noindent\textbf{Results.} Tab.~\ref{tab:all} shows that IDPro achieves new state-of-the-art performance~(89.6\% $\mathcal {J\&F}@60$) when across-frame scribbles are available. To ensure fairness, our IDPro was trained under a single-frame scenario when we compared it to the previous single-frame input methods. In a word, IDPro can achieve the SOTA performance regardless of the setting of single-frame/multi-frame and single-round/multi-round. 

From the results in Tab.~\ref{tab:all}, we find that: 
1) Despite notable improvements in inference speed, IDPro consistently outperforms the state-of-the-art model. In the 'old setting', compared to the MiVOS~(85.5 $J\&F@60$ in the first round and 88.5 $J\&F@60$ in the third round), our SwinB-IDPro demonstrates superior performance~(86.2 $J\&F@60$ in the first round and 88.7 $J\&F@60$ in the third round). 2) In the 'new setting', SwinB-IDPro achieves peak performance~(87.5 $J\&F@60$ in the first round and 89.6 $J\&F@60$ in the third round). Notably, IDPro is the pioneering model capable of assimilating multi-frame interactive information within a single iteration. This feature not only enriches user interaction but also bolsters processing speed and model efficacy after a single iteration. 3) Results from multiple iterations surpass those from a single iteration. This can be attributed to users iteratively refining segmentation via scribbles or click-based interactions before each iteration. More interactions lead to segmentation aligning more closely with user-defined objectives.

\begin{table}[t]
\centering
\caption{{\bf The performance on different input frame numbers and interaction rounds~(\S\ref{sec:ablation})}. $f$ denotes the number of input frames and $r$ denotes the number of interaction rounds.}
\begin{tabular}{c|cc|c|cc}
\toprule
\multirow{2}{*}{R50-IDPro} & \multicolumn{2}{c|}{$\mathcal {J\&F}@60$}      & \multirow{2}{*}{SwinB-IDPro} & \multicolumn{2}{c}{$\mathcal {J\&F}@60$}       \\ \cline{2-3} \cline{5-6} 
 & \multicolumn{1}{c|}{$r=1$} & $r=3$ &                  & \multicolumn{1}{c|}{$r=1$} & $r=3$ \\ \hline
$f=1$  & \multicolumn{1}{c|}{85.4} &  87.3   & $f=1$           & \multicolumn{1}{c|}{86.2} & 88.7  \\ 
$f=2$  & \multicolumn{1}{c|}{86.2} &  87.9   & $f=2$           & \multicolumn{1}{c|}{87.2}    &  89.0   \\ 
$f=3$  & \multicolumn{1}{c|}{87.0}    & 89.1 & $f=3$           & \multicolumn{1}{c|}{87.5}    & 89.6    \\ \bottomrule
\end{tabular}
\vspace{-0.3cm}
\label{tab:frame}
\end{table}

\begin{table}[t]
\centering
\caption{{\bf The memory and the time analysis for each sub-component~(\S\ref{sec:memory})}. For each sub-component in the propagation module at one round and single frame with 480p input size, we test the memory and time consumption.}
\begin{tabular}{ccc}
\toprule
R50-IDPro & Memory(MB) & Time \\ \midrule
AFI w/o AFA        & 3347       & 37 \\
AFA        & 354       & 5 \\
Memory Generate Network        & 3159       & 29.3 \\
ID-queried Decoder       & 3249       & 27.7 \\ \bottomrule
\end{tabular}
\vspace{-0.3cm}
\label{tab:conponent}
\end{table}

\begin{table}[t]
\centering
\caption{{\bf The memory and time consumption vary with different input sizes~(\S\ref{sec:memory})}. we test the memory and time consumption for AFA with different input sizes of frames.}
\begin{tabular}{c|cc|cc}
\toprule
\multirow{2}{*}{\makecell{input\\ size}} & \multicolumn{2}{c|}{Memory(MB)} & \multicolumn{2}{c}{Time(ms)} \\
                       & AFI w/o AFA          & AFA          & AFI w/o AFA        & AFA         \\ \midrule
360p                   & 3231             & 232          & 36.8           & 3.3         \\
480p                   & 3347             & 354          & 37.0           & 5.0         \\
600p                   & 3577             & 758          & 39.2           & 12.3        \\
720p                   & 4039             & 1188         & 56.1           & 15.5        \\ \bottomrule
\end{tabular}
\label{tab:input1}
\end{table}

\begin{table}[t]
\centering
\caption{{\bf The memory and time consumption vary with the number of input frames~(\S\ref{sec:memory})}. We test the memory and time consumption at 480p which varies with the number of input frames. AFA denotes the Across-Frame Attention in Sec~\ref{sec:s2m}, and we observe that AFA loses little efficiency when increasing the frame number.}
\begin{tabular}{c|cc|cc}
\toprule
\multirow{2}{*}{\makecell{Frame\\ number}} & \multicolumn{2}{c|}{Memory(MB)} & \multicolumn{2}{c}{Time(ms)} \\
                       & AFI w/o AFA          & AFA          & AFI w/o AFA        & AFA         \\ \midrule
$f=1$                   & 3347             & 354          & 37.0           & 5.0         \\
$f=2$                   & 3551             & 502          & 53.9           & 12.7        \\
$f=3$                   & 3709             & 784          & 77.6           & 22.2        \\
$f=4$                   & 3881             & 1090         & 105.3           & 23.6        \\ 
$f=5$                   & 4083             & 1336         & 119.1           & 38.6        \\ \bottomrule
\end{tabular}
\label{tab:input}
\end{table}

\begin{table}[t]
\centering
\caption{{\bf Sensitivity Study of Scribbles~(\S\ref{sec:ablation}).} We divided scribbles into three qualities: detailed, general, and rough to verify their performance separately.}
\begin{tabular}{c|c|c}
\toprule
 Scribble & MiVOS & IDPro  \\ \midrule
 Detailed & 85.9 & 86.3 \\ 
 General  & 85.1 & 85.2 \\ 
 Rough  & 84.8 & 85.0 \\ \bottomrule
\end{tabular}
\vspace{-0.2cm}
\end{table}

\begin{figure*}[t]
    \centering
\includegraphics[width=0.95\textwidth]{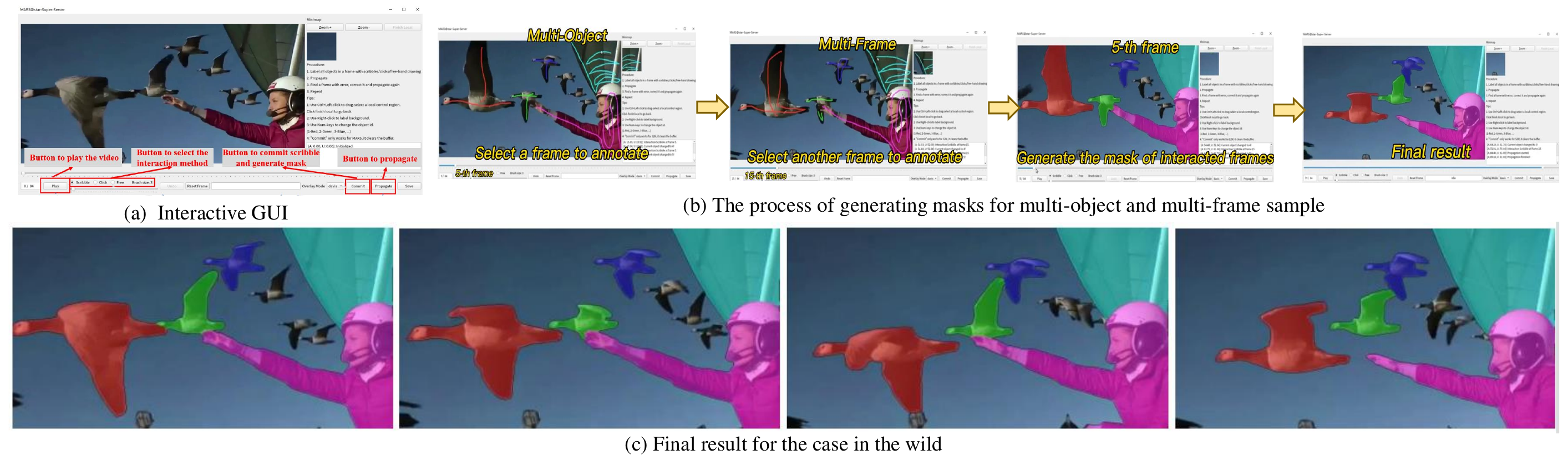}
    \vspace{-0.3cm}
    \caption{{\bf The interactive graphical user interface~(GUI; \S\ref{sec:GUI})}. In~(a), a succinct GUI streamlines user operations, enhancing the user experience. In~(b), IDPro entails the creation of masks for multi-object and multi-frame datasets. In (c), our model exhibits commendable efficacy even in the case of the wild. }
    \vspace{-0.3cm}
    \label{fig:visil_}
\end{figure*}

\subsection{Ablation Study}
\label{sec:ablation}
\noindent\textbf{Settings.} Our baseline is designed to process multi-frame interaction data, incorporating a cross-frame interaction module~(AFI) devoid of cross-frame attention~(AFA) while accommodating multi-frame inputs. Our baseline was trained on the LVIS dataset and adopted the single-propagation strategy akin to Mivos~\cite{ChengTT21} during the propagation phase. Our ablation study primarily investigates the impact of the cross-frame attention~(AFA) module, the re-propagation strategy, and additional training on the YTB \& DAVIS dataset.
\\~

\noindent\textbf{Results.}
From the results in Tab.~\ref{tab:ablation} and Tab.~\ref{tab:frame}, we find that:

\textit{1) Effectiveness of Across-Frame Interaction.} 
The Across-Frame Interaction Module~(AFI) can accept multiple frames and generate their masks. Moreover, the AFI incorporates an Across-Frame Attention~(AFA) that effectively aggregates multi-scale semantic information and captures temporal information across the interacted frames~(see Fig.~\ref{fig:compare}). Tab.~\ref{tab:ablation} demonstrates that multi-scale semantic and temporal information can improve performance.

\textit{2) Effectiveness of Concurrent Propagation.}  
Multi-round interactions have been demonstrated to outperform single-round interactions because each interaction round contains more user intent. Thus, we design the across-round memory to store important interaction information and re-propagate it for all frames without interaction. Tab.~\ref{tab:ablation} shows the benefit of this propagation strategy.

\textit{3) Performance Varies with Frame Number and Interaction Rounds.} In Tab.~\ref{tab:frame}, as the number of interacting frames increases within the same iteration round, users can furnish more guiding information, consequently enhancing IDPro performance. For instance, SwinB-IDPro demonstrates 86.2 at $J\&F$ in a single round with single frames, escalating to $87.5$ at $J\&F$ in a round involving three frames. Hence, our proposed IDPro, facilitating multiple user interactions within a single round, proves indispensable. This approach not only grants users greater interaction flexibility but also enhances accuracy within identical iteration rounds. Moreover, as the iteration count rises, meticulous tuning of each iteration's rounds based on previous segmentation augments the model's performance. 

\lkx{\textit{4) Performance Varies with the level of detail scribbles.}We divided scribbles into three qualities: detailed, general, and rough following. Specifically, IDPro achieves 86.3, 85.2~(-1.1), 85.0~(-1.3)~($\mathcal {J\&F}@60$) on the three scribble types respectively. Compared to SOTA, which achieves 85.9, 85.1~(-0.8), and 84.8~(-1.1). Experimental results yield that the level of detail in user-provided scribbles significantly impacts performance. Specifically, more intricacy within the scribbles corresponds to improved output quality. Moreover, the consistent efficacy of IDPro across varying degrees of meticulous scribbling. These observations underscore the pivotal role of scribble precision in influencing outcomes and emphasize the stability of IDPro performance.}

\begin{figure*}[t]
    \centering
    \includegraphics[width=1\textwidth]{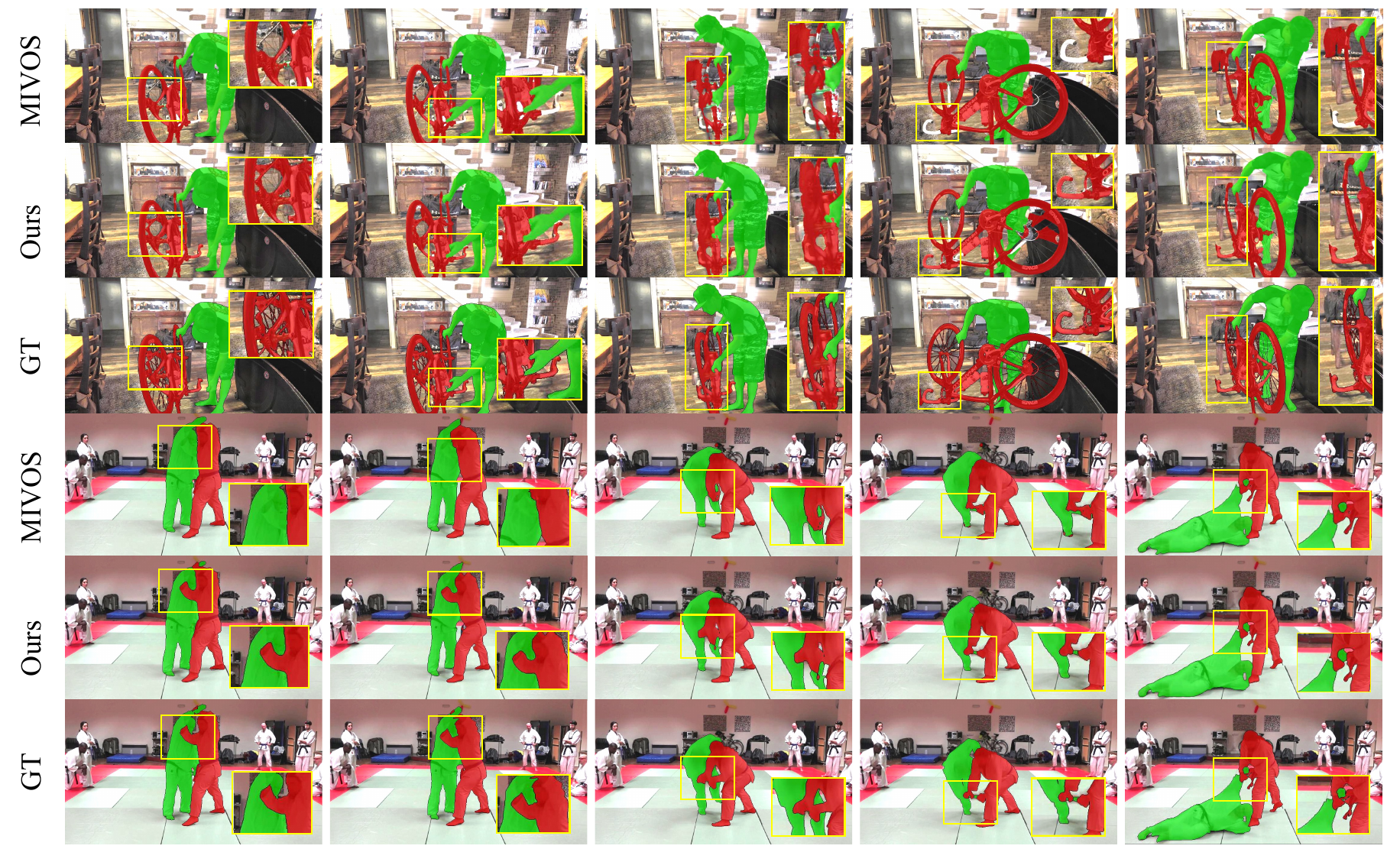}
    \caption{{\bf Qualitative comparison on the DAVIS interactive track~(\S\ref{sec:acc})}. We compare our method IDPro with MiVOS~\cite{ChengTT21} on the DAVIS dataset.}
    \label{fig:visil}
    \vspace{-0.3cm}
\end{figure*}

\subsection{Memory and Time Consumed Analysis}
\label{sec:memory}
The proposed across-frame attention~(AFA) boosts model performance by assimilating and leveraging information across frames. However, in models necessitating real-time interaction, memory usage and inference speed become pivotal factors. Hence, we delved into the alterations in memory usage and inference speed linked to changes in the number of frames and input size attributed to the across-frame attention~(AFA) and the across-frame interaction module~(AFI).

\textit{ 1) Frame Number.}
The Across-Frame Interaction Module~(AFI) functions to derive segmentation results from user-provided scribbles. As the number of interacting frames rises, the model naturally incurs higher memory usage and inference time. Yet, our analysis from Tab.~\ref{tab:input} reveals a controlled increase in our across-frame attention with expanding frame numbers. Notably, our across-frame interaction module loses little efficiency when increasing the frame number. This highlights the viability of our proposed method for multi-frame interaction within a single iteration, indicating no significant strain on memory or processing speed.

\textit{ 2) Input Size.}
The memory usage and inference speed are intricately tied to the size of input frames. Consequently, we scrutinized how alterations in input frame size impacted the memory usage and inference speed of both the Across-Frame Interaction Module~(AFI) omits the Across-Frame Attention~(AFA) and AFA. Analysis from Tab.~\ref{tab:input1} highlights a discernible trend: our pivotal across-frame attention, in contrast to the comprehensive interaction module, incurs minimal memory consumption and negligible computational time.

\textit{ 3) Memory and Time Consumed for Each Subcomponent.}
Tab.~\ref{tab:conponent} illustrates the memory and time demands of submodules. Notably, within the concurrent propagation module, MU-Net and MP-Net dominate memory usage. In contrast, our proposed multi-frame interaction module maintains a manageable memory. Moreover, the propagation module necessitates more inference time compared to the interaction module. Overall, IDPro exhibits commendable and controlled performance in both memory usage and inference time.

\subsection{Interactive Graphical User Interface}
\label{sec:GUI}
We developed an interactive graphical user interface~(GUI) that enables multi-frame interaction for mask generation. In Fig.~\ref{fig:visil_}~(a), a succinct visual interactive system streamlines user operations, enhancing the user experience. The graphical user interface features intuitive buttons for sketching annotations and generating corresponding frame segmentation results. Additionally, a dedicated button facilitates the propagation of segmentation results from interacted frames to the entirety of the sequence. Moving to Fig.~\ref{fig:visil_}~(b), the procedural framework entails the creation of masks for multi-object and multi-frame datasets. Initially, users designate a frame for annotating scribbles, thereby enabling the annotation of multiple objects within the frame. Subsequently, users have the option to extend their annotations to additional frames. Upon completion, activating the ``commit scribbles'' button produces segmentation results encompassing all annotated frames. A concluding step involves users clicking the ``propagate'' button, yielding comprehensive segmentation results for the entire video. In Fig.~\ref{fig:visil_}~(c), our model exhibits commendable efficacy even in the case of the wild.

Using the interactive GUI tool, users can annotate multiple objects on multiple frames, and the IDPro automatically generates masks for all frames concurrently. If the user finds the generated mask to be unsatisfactory, another round of interaction can be carried out to improve it. IDPro then propagates the refined masks to all frames in the video, resulting in the segmentation of the entire video. IDPro enables efficient and accurate mask generation while reducing the user's workload and time spent on propagation.

\section{Conclusion}
iVOS is a human-computer interaction task that requires the quality of user experience, but previous methods are limited to single input mode and slow running speed. Thus, we propose IDPro, a novel multi-frame and multi-object framework that allows the users to interact with multiple frames at once, which is a pioneer of multi-frame interaction. In detail, IDPro achieves new SOTA performance on DAVIS 2017 dataset~(89.6\% $\mathcal {J\&F}@60$) when across-frame scribbles are available. Meanwhile, our R50-IDPro is more than $\bf{3 \times}$ faster than the SOTA competitor under challenging multi-object scenarios. Concretely, we proposed an across-frame interaction module that facilitates the transfer of scribble-based information between interactive frames. We develop a concurrent propagation module that can simultaneously process multiple objects, and design a re-propagation strategy to exploit the interactive information from multiple rounds. Additionally, we introduced an interactive graphical user interface~(GUI) that supports multi-frame interaction. We hope our IDPro can serve as a solid multi-frame and multi-object baseline in iVOS and will promote the use of iVOS in real-world applications.

\noindent{\bf Broader Impact and Limitations.}
1) The efficacy of iVOS methods is largely contingent on input scribble quality. In cases where the input scribbles are rough, segmentation performance may suffer.  2) IDPro is meticulously engineered to concurrently process multiple frames and objects, presenting an optimal solution for forthcoming tracking requirements aimed at minimizing time consumption.

\section*{Acknowledgments}
This work was supported by the National Key Research \& Development Project of China (2021ZD0110700), the National Natural Science Foundation of China (62337001), the Fundamental Research Funds for the Central Universities(226202400058) and \lk{the Fundamental Research Funds for the Central Universities (No. 226-2022-00051).}

\bibliographystyle{IEEEtran}
\bibliography{sample-base}

\begin{IEEEbiography}[{\includegraphics[width=1in,height=1.15in, clip,keepaspectratio]{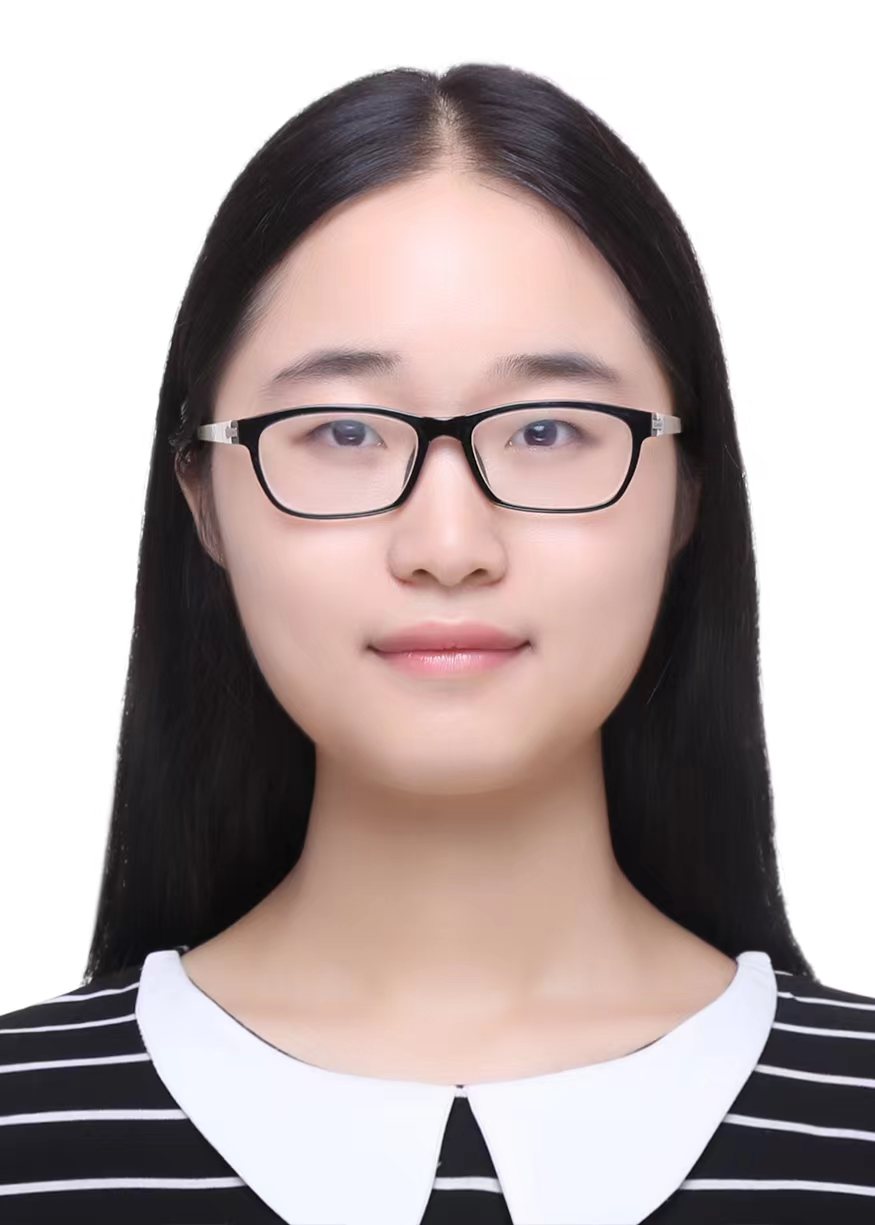}}]{Kexin Li} is currently a Ph.D. student in the College of Computer Science at Zhejiang University, Hangzhou, China. She received her B.S. degree from the Northeast Normal University, Changchun, China, in 2020. Her
current research interests include multimedia and video segmentation.
\end{IEEEbiography}

\begin{IEEEbiography}[{\includegraphics[width=1in,height=1.15in, clip,keepaspectratio]{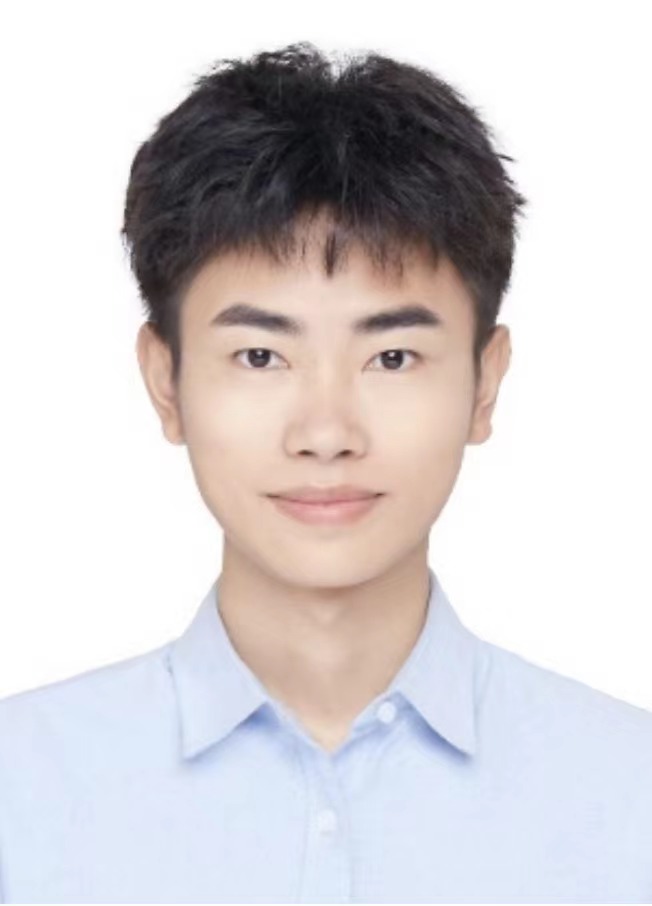}}]{Tao Jiang} received the B.E. degree from the School of Information Engineering, Zhejiang University of Technology, in 2021. He is currently pursuing a master’s degree with the Colledge of Software, at Zhejiang University. His research interests include computer vision and multimodal.
\end{IEEEbiography}

\begin{IEEEbiography}[{\includegraphics[width=1in,height=1.25in,clip,keepaspectratio]{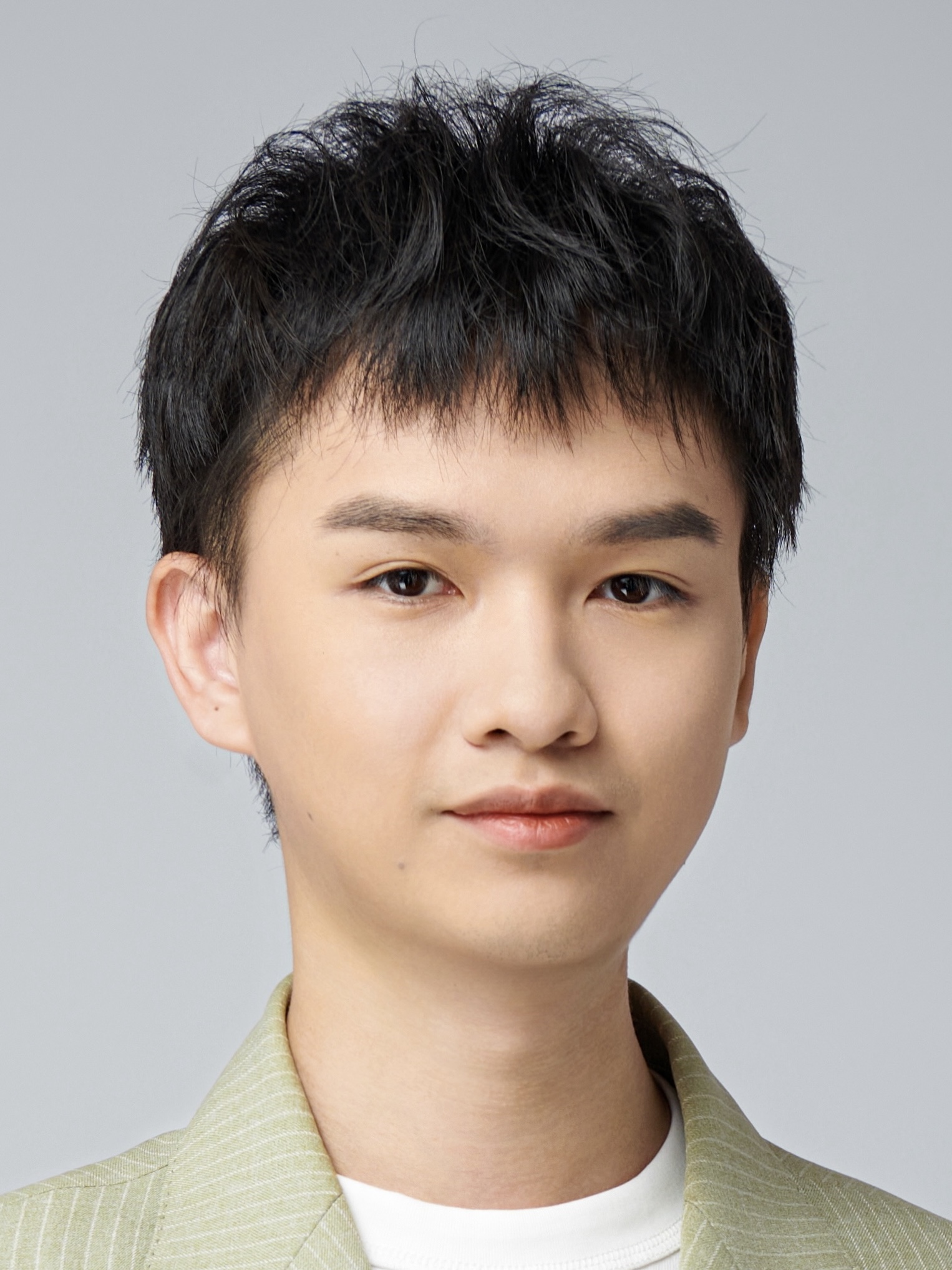}}]{Zongxin Yang} received his BE degree from the University of Science and Technology of China in 2018, and his PhD in computer science from the University of Technology Sydney, Australia, in 2021. He is currently a research fellow at Harvard University, USA, and was previously a postdoctoral researcher at Zhejiang University, China. His current research interests include multi-modal learning and its applications in biomedicine.
\end{IEEEbiography}

\begin{IEEEbiography}[{\includegraphics[width=1in,height=1.25in,clip,keepaspectratio]{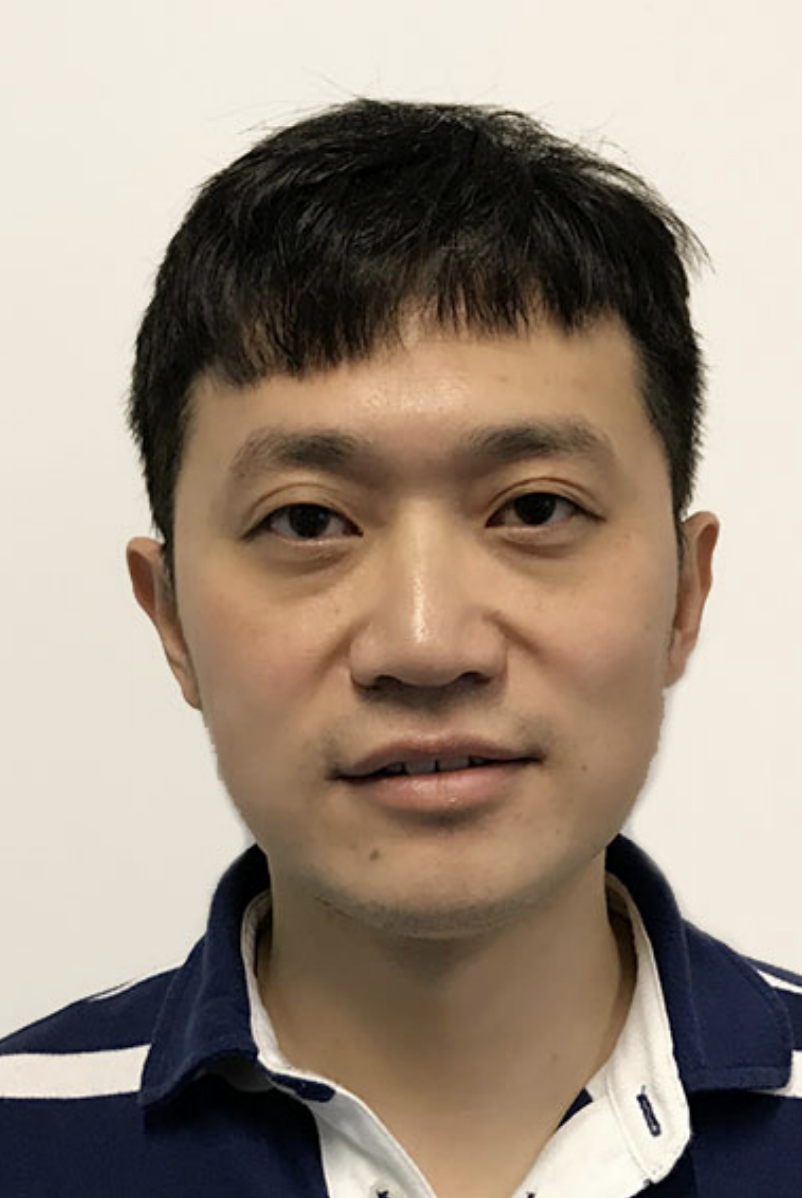}}]{Yi Yang} (Senior Member, IEEE) received the PhD
degree from Zhejiang University, in 2010. He is
a distinguished professor at Zhejiang University, China. His current research interests include machine learning and multimedia content analysis, such as multimedia
retrieval and video content understanding. He received the Australia Research Council Early Career Researcher Award, the Australia Computing Society, the Google Faculty Research Award, and the ACS Machine Learning Research Award Gold Disruptor Award.
\end{IEEEbiography}

\begin{IEEEbiography}[{\includegraphics[width=1in,height=1.25in,clip,keepaspectratio]{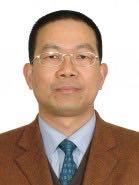}}]{Yueting Zhuang}
received his B.Sc., M.Sc. and Ph.D. degrees in Computer Science from Zhejiang University, China, in 1986, 1989 and 1998 respectively. From February 1997 to August 1998, he was a visiting scholar at the University of Illinois at Urbana-Champaign. He served as the Dean of College of Computer Science, Zhejiang University from 2009 to 2017, the director of Institute of Artificial Intelligence from 2006 to 2015. 
He is now a CAAI Fellow~(2018) and serves as a standing committee member of CAAI.  He is a Fellow of the China Society of Image and Graphics~(2019).  Also, he is a member of Zhejiang Provincial Government AI Development Committee~(AI Top 30).
\end{IEEEbiography}

\begin{IEEEbiography}[{\includegraphics[width=1in,height=1.25in,clip,keepaspectratio]{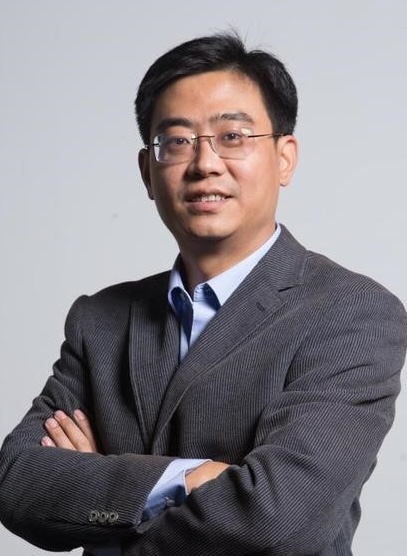}}]{Jun Xiao}
received a Ph.D. degree in computer science and technology from the College of Computer Science, Zhejiang University, Hangzhou, China, in 2007. He is currently a professor at the College of Computer Science, Zhejiang University. His current research interests include computer animation, multimedia retrieval, and machine learning.
\end{IEEEbiography}

\vfill

\end{document}